\documentclass{article}
\usepackage{spconf,amsmath,graphicx}
\usepackage{multirow}
\usepackage[table]{xcolor}
\usepackage{csquotes}
\usepackage{tabularx}
\usepackage{algorithm}
\usepackage{algpseudocode}
\usepackage{bm}

\title{CTC-GMM: CTC guided modality matching for fast and accurate streaming speech translation}
\name{Rui Zhao, Jinyu Li, Ruchao Fan, Matt Post}
\address{Microsoft, USA\\
}
\begin{document}
%
\maketitle
%
\begin{abstract}
Models for streaming speech translation (ST) can achieve high accuracy and low latency if they're developed with vast amounts of paired audio in the source language and written text in the target language. Yet, these text labels for the target language are often pseudo labels due to the prohibitive cost of manual ST data labeling. In this paper, we introduce a methodology named Connectionist Temporal Classification guided modality matching (CTC-GMM) that enhances the streaming ST model by leveraging extensive machine translation (MT) text data. This technique employs CTC to compress the speech sequence into a compact embedding sequence that matches the corresponding text sequence, allowing us to utilize matched {source-target} language text pairs from the MT corpora to refine the streaming ST model further. Our evaluations with FLEURS and CoVoST2 show that the CTC-GMM approach can increase translation accuracy relatively by 13.9\% and 6.4\% respectively, while also boosting decoding speed by 59.7\% on GPU.  
\end{abstract}

\section{Introduction}
\label{sec:intro}

Speech translation (ST) targets on the translation of spoken audio in one language into text in another language. Most recent ST systems use end-to-end (E2E) models \cite{sperber2020speech}, which are different from the traditional cascaded systems that translate automatic speech recognition (ASR) models' output with the machine translation (MT) models. The most common E2E ST models \cite{Berard2016ST, weiss2017sequence, vila2018end, radford2023robust} are attention-based encoder-decoder (AED) models \cite{cho2014learning, Attention-bahdanau2014}.

Streaming simultaneous speech translation (ST) \cite{ma2021streaming}, which translates speech as it is spoken, is a crucial topic in the ST area. To stream AED models, chunkwise attention is needed, such as MoChA \cite{chiu2018monotonic}, MILk \cite{arivazhagan2019monotonic} and monotonic multi-head attention \cite{ma2019monotonic, ma2021streaming}. The most recent representation of streaming AED ST model is SeamlessM4T v2 \cite{barrault2023seamless}. 

Streaming is also the most requested feature in ASR \cite{sainath2020streaming, Li2020Developing}. The streaming AED methods have been shown to be less effective than the recurrent neural network Transducer (RNN-T) \cite{Graves-RNNSeqTransduction}, which is now the most common streaming E2E ASR model \cite{li2022recent, prabhavalkar2023end}. Building on the success of streaming E2E ASR models, RNN-T based E2E models were developed for low-latency and high-quality ST \cite{xue2022large, xue2023weakly}. Although RNN-T models are monotonic, they can deal with the word-reordering challenge in ST by using the flexible RNN-T path during decoding as described in \cite{xue2022large}. Those models are reported to be used in commercial products \cite{3pAPI}, even without the need of cloud connection \cite{PCtranslation}.     

To develop high-quality E2E ST models, we usually need a large amount of paired speech and text data. However, it is much more expensive to use humans to label ST data than ASR data. Therefore, a common practice is to use a MT model to generate pseudo labels of target language from the reference texts in source-language ASR corpus \cite{jia2019st,gaido2020end}. These pseudo labeled data are then used to train E2E ST models. However, pseudo labels have errors that may affect the performance of E2E ST models. To further improve ST model quality, we should also use the paired source/target language text corpus that are used for training MT models.  

In this paper, we propose CTC guided modality matching (CTC-GMM) to improve the RNN-T based streaming ST model by using MT text training data. This is achieved by using Connectionist Temporal Classification (CTC) \cite{graves2006connectionist} to transform the speech sequence into a shorter embedding sequence that resembles the text sequence. Then the speech embedding sequence or the text sequence is fed into a shared encoder. In this way, the matched {source-target} text pairs in MT corpus can be used along with matched speech-text data to build the streaming ST model. Due to the fact that the speech embedding sequence is much shorter than the original speech sequence, the cost of the shared encoder inference and the decoding steps can be reduced, making the proposed model faster and more accurate. We evaluate the proposed method for translating German audio into English text with FLEURS \cite{fleurs2022arxiv} and CoVoST2 \cite{wang2020covost} test sets. Our proposed CTC-GMM model achieves a relative improvement of 13.9\% and 6.4\% respectively in BLEU scores compared to the baseline ST model.  

The structure of this paper is as follows: Section \ref{sec:related} outlines the related work. In Section \ref{sec:CTCGMM}, we detail our CTC-GMM method. Section \ref{sec:exp} describes the experimental setup, while Section \ref{sec:res} presents the findings. We conclude the paper in Section \ref{sec:con}.

\section{Related Works}
\label{sec:related}

The authors of \cite{gaido2021ctc} introduced CTC compression to reduce the length of the speech sequence based on its phonetic features. Our CTC-GMM work instead applies CTC to match the speech embedding and text sequence so that MT training data can be used. Moreover, to overcome the discrepancy between the actual labels in training and inferred labels in testing, we propose a sampling strategy inspired by \cite{ruchao2023sampling}, which is shown to be essential for our algorithm.

Many works have been done on joint speech-text modeling. SpeechT5 \cite{ao2022speecht5} and speechLM \cite{zhang2024speechlm} align the speech and text modalities with mix-up or swapping strategy. 
In \cite{barrault2023seamless}, length adaptor is applied to minimize the length discrepancy between speech and text, similar to M-adaptor \cite{zhao2022Madaptor} which uses multi-head pooled self attention.  
Prior works also heavily rely on pre-training for speech and text modality alignment. In contrast, our proposed CTC-GMM does not need any pre-training and directly learn the alignment through CTC targets on top of the speech encoder. 

All the related works mentioned above are based on either AED or CTC models. Maestro \cite{chen22MAESTRO} learns shared representation for speech and text in RNN-T framework to benefit ASR. However, it requires duration modeling to upsample the text sequence to match the speech sequence length. Neither the shared encoder inference nor the decoding steps are reduced in terms of the computational costs. 

One approach to enhance ST models with MT data involves using text-to-speech (TTS) systems to generate audio from the source text in paired MT data \cite{jia2019st}. The resulting source audio and target text pair is then used for training ST models. However, this method is expensive as it requires a separate TTS system and careful handling, like freezing the speech encoder, to prevent bias towards TTS speakers \cite{zhao2021addressing}. Although the recent advancement of zero-shot TTS \cite{wang2023neural, borsos2023soundstorm, ju2024naturalspeech, le2024voicebox} can enrich the speaker diversity of generated audio, it needs even larger cost.  In contrast, our proposed method doesn't have those constraints, directly consuming the text data within the model. 

One benefit of our proposed method is to significantly reduce the decoding time. There are methods using uniform or adaptive downsampling to reduce the frame rate for fast computation \cite{burchi2021efficient, li23Accelerating, prabhavalkar2024extreme}. However, these methods cannot be used to leverage text training data for model quality improvement. 

\section{Streaming speech translation with CTC guided modality matching}
\label{sec:CTCGMM}

In this section, we will first introduce streaming speech translation with RNN-T models, and then describe the proposed CTC guided modality matching method in detail. 

\subsection{Streaming speech translation with RNN-T}
\label{ssec:rnnt}

\begin{figure}[t]
    \centering
    \includegraphics[width=0.36\textwidth]{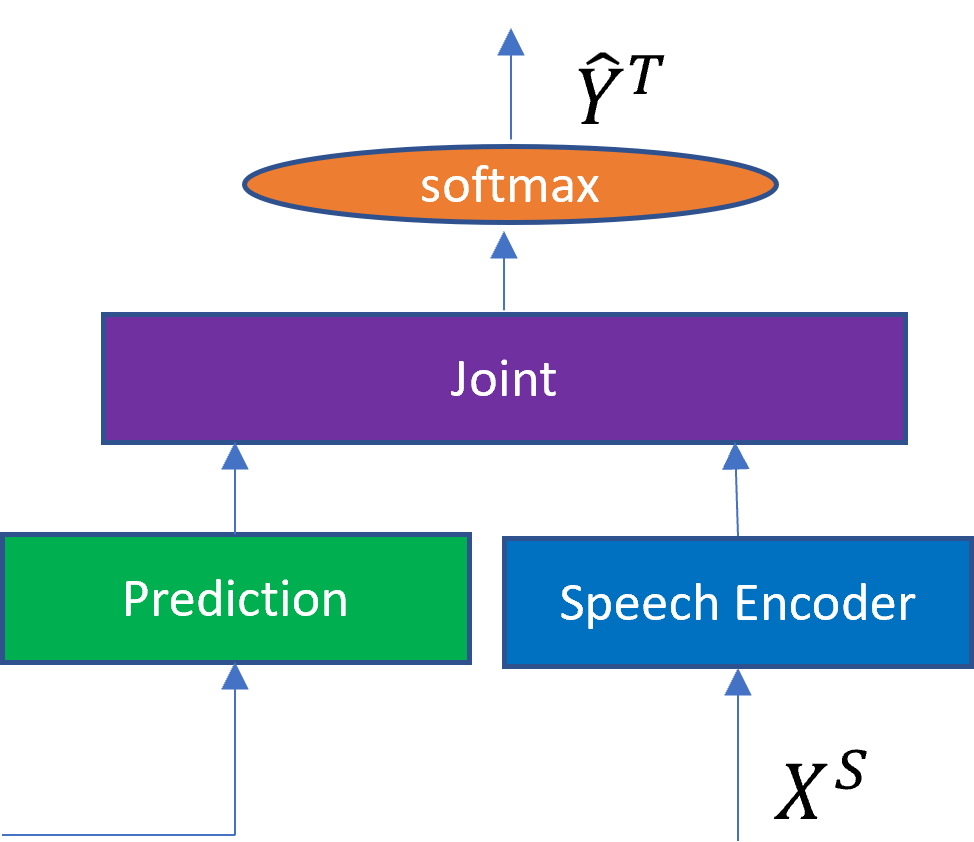}
    \caption{The RNN-T Structure.}
    \label{fig:RNNT}
\end{figure}

Compared to ASR, it is much more expensive to obtain human-labeled data for ST. A common practice is to use a MT model to generate pseudo labels from the reference texts in ASR corpus \cite{jia2019st,gaido2020end} and those pseudo labeled data are then used to train E2E ST models. The source speech and transcription are represented as $X^S$ and $Y^S$, respectively. After feeding $Y^S$ into a MT system, the pseudo label in target language is represented as $\hat{Y}^T$. Here, $S$ and $T$ denote the source and target languages, respectively.  

The baseline streaming speech translation model with RNN-T structure is shown in Figure \ref{fig:RNNT}. The speech sequence $X^S$ in the source language is mapped to the text sequence $\hat{Y}^T$ in the target language. The speech encoder produces high-level feature from $X^S$, and the prediction network produces predictor feature from previous non-blank output token. Then, a joint network combines the outputs of the speech encoder and prediction network, and generates the output translated text.

RNN-T inference is a frame-synchronized decoding with beam search. Therefore, the decoding cost is proportional to the sequence length of the speech encoder output. The standard RNN-T beam search decoding algorithm is shown in Algorithm 1. Here $\Pr(\bm{y})$ is the approximate probability of emitting output sequence $\bm{y}$ found by the search so far. 
$\Pr(k|\bm{y}, t)$ is the probability of extending $\bm{y}$ by
$k\in\bar{\mathcal{Y}}$ at time $t$, where $\bar{\mathcal{Y}}$ is the extended output space set including all output label set $\mathcal{Y}$ and blank $\emptyset$. $\textit{pref}(\bm{y})$ is the set of proper prefixes of $\bm{y}$, and for $\hat{y}\in\textit{pref}(\bm{y})$, let $\Pr(\bm{y}|\hat{\bm{y}}, t) = \prod_{u=|\hat{\bm{y}}|+1}^{|\bm{y}|}\Pr(y_u|\bm{y}_{[0:u-1]}, t)$. $W$ is the beam size. The probability $\Pr(k|\bm{y}, t)$ can be written as:
\begin{eqnarray}
 \Pr(k|\bm{y}, t) = \text{softmax}(z_{t, \bm{y}}) \nonumber \\
 z_{t, \bm{y}} = Joint(h_{enc}^t, h_{dec}^{\bm{y}})
 \label{eqn:rnnt}
\end{eqnarray}

As shown in Algorithm \ref{alg:RNNT}, for each time $t$, given the encoder output $h_{enc}^t$, we need to generate new $W$ candidates set $B$ based on $W$ candidates set $A$ from $t-1$. For each candidate $\bm{y}^*$ from A, the output probability for $k\in\bar{\mathcal{Y}}$ given $\bm{y}^*$ need to be calculated, which involves the inference of prediction and joint network to get $h_{dec}^{\bm{y}}$ and $z_{t, \bm{y}}$ respectively (as shown in equation \ref{eqn:rnnt}), as well as $|\bar{\mathcal{Y}}|$ loops to do the expansions. Hence the decoding time could be decreased significantly if we reduce the number of frames of the encoder output.  

\begin{algorithm}
\caption{RNN-T Beam Search decoding}
\begin{algorithmic}
\State \textbf{Initialize}: $B = \{\emptyset\} ; \Pr(\emptyset) = 1$

\For{$t = 1$ to $T$}
    \State $A = B$
    \State $B = \{\}$
    \For{$ \bm{y} \in A$}
        \State $\Pr(\bm{y}) \mathrel{+}= \sum_{\hat{\bm{y}} \in \text{pref}(\bm{y}) \cap A} \Pr(\hat{\bm{y}}) \Pr(\bm{y}|\hat{\bm{y}}, t)$
    \EndFor
    \While{$B$ contains less than $W$ elements more probable than the most probable in $A$}
        \State $ \bm{y}^* = $ most probable in $A$
        \State Remove $\bm{y}^*$ from $A$
        \State $\Pr(\bm{y}^*) = \Pr(\bm{y}^*) \Pr(\emptyset | \bm{y}^*, t)$
        \State Add $\bm{y}^*$ to $B$
        \For{$k \in \mathcal{Y}$}
            \State $\Pr(\bm{y}^* + k) = \Pr(\bm{y}^*) \Pr(k | \bm{y}^*, t)$
            \State Add $\bm{y}^* + k$ to $A$
        \EndFor
    \EndWhile
    \State Remove all but the $W$ most probable from $B$
\EndFor
\State \textbf{Return:} $y$ with highest $\log \Pr(y)/|y|$ in $B$
\end{algorithmic}
\label{alg:RNNT}
\end{algorithm}

\subsection{CTC guided modality matching}
\label{ssec:method}

\begin{figure}[t]
    \centering
    \includegraphics[width=0.5\textwidth]{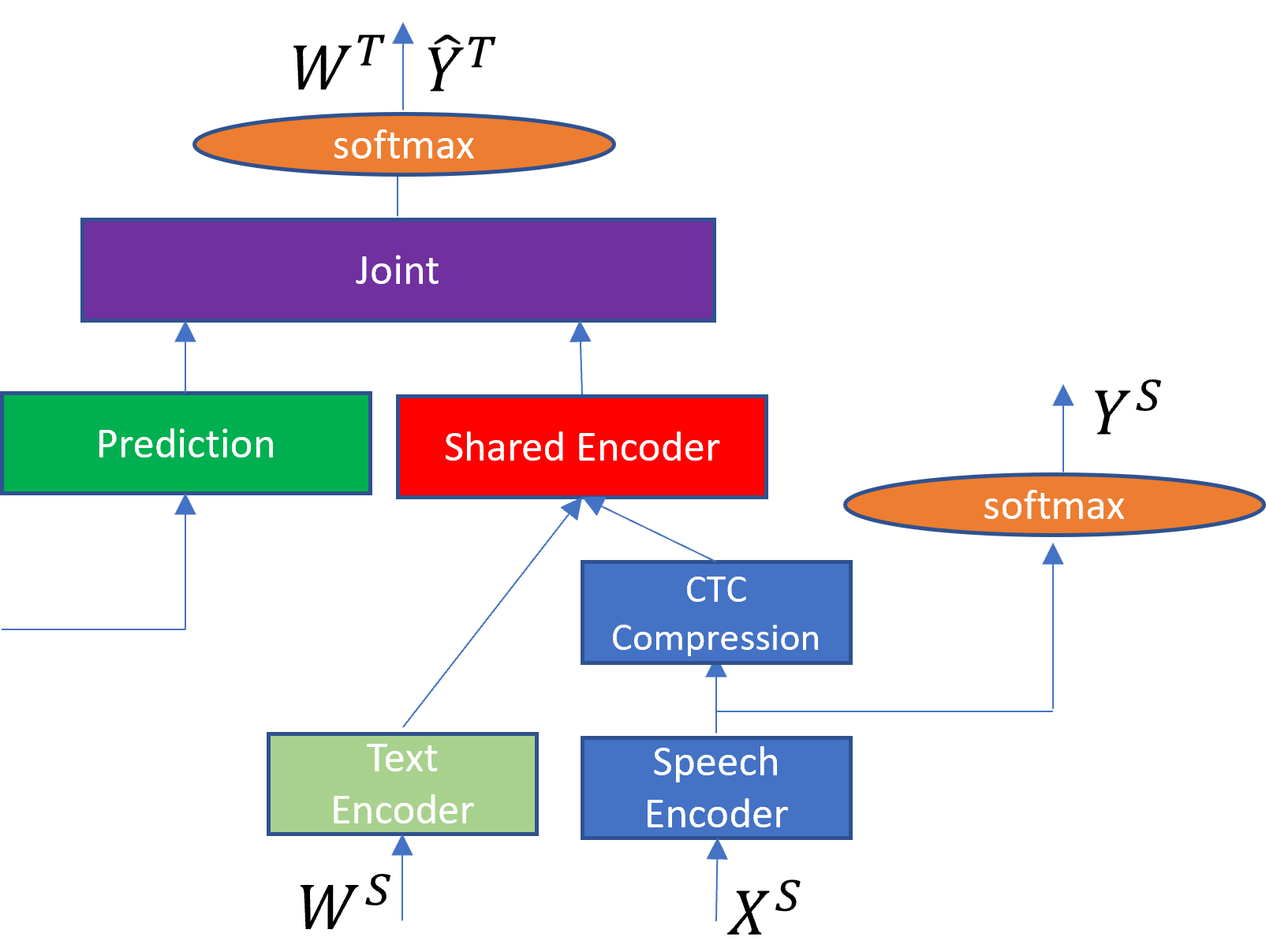}
    \caption{The CTC-GMM structure}
    \label{fig:CTCGMM}
\end{figure}

In Figure \ref{fig:CTCGMM}, we show the proposed CTC guided modality matching (CTC-GMM) based on the RNN-T structure, which enables the usage of MT data to further improve the ST quality. The text of source and target in the MT corpus are denoted as $W^S$ and $W^T$, respectively.  The source audio, source transcription, and target pseudo transcription in the speech corpus are $X^S$, $Y^S$, and $\hat{Y}^T$ respectively. Note that the source text in the MT and source transcription in speech corpus are different, denoted as $W^S$ and $Y^S$, respectively. However, the words of them are converted into the same set of BPE tokens using the same dictionary to ensure better alignment the text and speech modalities. The model takes either the speech input $X^S$  or the text input $W^S$. 

The speech encoder consumes $X^S$ as the input and predicts the ASR label $Y^S$ using the CTC loss. The speech encoder output sequence is denoted as $\{\bm{h}_1, \bm{h}_2, ......,  \bm{h}_L\}$, where $L$ is the sequence length. The text encoder is a text embedding layer with the same output dimension as the speech encoder output.  After linear transform and softmax operation, the corresponding output CTC probability vector is $\{\bm{o}_1, \bm{o}_2, ......,  \bm{o}_L\}$. 

Then the CTC compression module squeezes the encoder output based on the CTC probability in two steps. In the first step, the predicted token index for each time $l$ is obtained by either selecting the index  with the highest CTC probability or by sampling method similar as in \cite{ruchao2023sampling}. Specifically for the sampling-based method, for each CTC output $\bm{o}_l$ ($l \in [1,......,L]$), the indexes of the top $N$ largest output values are picked, denoted as $\{k_1$, $k_2$, ......, $k_{N}\}$ 
where $k_n \in\bar{\mathcal{Y}}$, and then we generate a rand number $r$ between $1$ and $N$ according to the below probabilities 
\begin{equation}
   prob(r=n) = \frac{o_{l,k_n}}{\sum_{i=1}^N{o_{l,k_i}}}   (n=1,2,......,N).
\end{equation}
$k_r$ is finally selected as the predicted token index of time $l$. $N$ is set to be 5 in the experimental settings. 

The second step is to merge the consecutive speech encoder output $\bm{h}_l$ with the same predicted tokens into one frame. Suppose the predicted token of time $l$ is $v_l$, the frames $l=\{i,....,j\}$ have the  same predicted tokens $v$. The merged encoder output $\hat{\bm{h}}$ could be calculated with several options:
\begin{itemize}
    \item Average: $\hat{\bm{h}} = \sum_{l=i}^j{\bm{h}_l} / (j-i+1)$ 
    \item Attention: $\hat{\bm{h}} = \textit{Att}(q^u, W^K h^j_{i^{'}}, W^V h^j_{i^{'}})$,
    \item Discrete: $\hat{\bm{h}} = \textit{Embedding}(v)$ 
\end{itemize}
where $W^K$ and $W^V$ represents the linear layers to obtain key and value in the attention layer, respectively. $q^u$ is the sinusoidal positional embedding with position $u$ (the $u^{th}$ merged output).
Both the Average and Attention options generate continuous embedding which can be directly hooked up with shared encoder.
For the attention option, the consecutive frames with blank as the predicted token are merged with the subsequent consecutive frames with non-blank labels as the predicted token. Hence the beginning frame index is $i^{'}$ instead of $i$, where the predicted token of frames $i^{'}$ to $i$ is blank. 
For the discrete option, the prediction token needs to go through an embedding layer to generate continuous embedding to hook up with the shared encoder. As CTC predicts the blank token in addition to the standard BPE tokens, we further have the option to remove all blank tokens in the discretized sequence.  

\begin{table}[tp]
    \centering
    \begin{tabular}{|c|c|}
    \hline
        Compression Options & $W^S$ processing \\ \hline \hline
        Average &  insert blank between bpe tokens \\ \hline
        Attention & generate bpe tokens \\  \hline
        Discrete &  insert blank between bpe tokens \\  \hline
        Discrete: blank removal &  generate bpe tokens \\  \hline
    \end{tabular}
    \caption{The processing of MT data input $W^S$ for different CTC compression options.}
    \label{tab:ws_processing}
\end{table}

CTC compression makes the input sequence shorter for both the shared encoder and joint network. This reduces the computation cost when the shared and speech encoders have the same number of layers as the baseline model's speech encoder. It also reduces the decoding time because it needs fewer decoding steps as explained in Section \ref{ssec:rnnt}.

To use text data from MT corpus in the ST model, we input the paired $\{W^S, W^T\}$ data into the model. Depending on the CTC compression operation we use, $W^S$ has to be modified as in Table \ref{tab:ws_processing}. For the Average or Discrete operation in CTC compression, both standard BPE tokens and blank token are predicted labels. Therefore, to make sure the text sequence of the input of MT corpus and the CTC compression output are matched, we need to insert blank labels between BPE tokens. On the other hand, the Attention operation and 
Discrete with blank removal operation already eliminate the predicted blank tokens, we only need to convert the word sequence into BPE tokens.

The final objective function is the sum of CTC ASR loss for $\{X^S, Y^S\}$, RNN-T ST loss for $\{X^S, \hat{Y}^T\}$, and RNN-T MT loss for $\{W^S, W^T\}$:
\begin{equation}
   \ninept L = 0.1 L_{CTC}(Y^S|X^S) + L_{RNN-T} (\hat{Y}^T|X^S)  + L_{RNN-T} (W^T|W^S).
\end{equation}
In every minibatch during training, the ratio of data from speech corpus and MT corpus is $1:1$. 

\section{Experimental Settings}
\label{sec:exp}

The experiments are to build ST models that  translate German audio into English text. The speech training data is from an in-house German ASR corpus with 30 thousand (K) hours speech data with the mixed bandwidth data \cite{li2012improving}. All the training data is anonymized with personally identifiable information removed. We use a text-based MT service to convert the German ASR transcriptions into English texts as the pseudo labels for ST training. The MT corpus has totally 280 million (M) paired German to English text sentences.

We use the German-to-English test sets in FLEURS and CoVoST2 to evaluate the performance of the proposed CTC-GMM model. The training speech and MT corpora do not contain the FLEURS and CoVoSR2 data.

The acoustic feature is 80-dimension log Mel filter bank for every 10 ms speech. The baseline E2E ST model is built on the conformer transducer structure\cite{gulati2020conformer} with a streaming setup \cite{chen2021developing}. The Convolution sub-sampling layer contains 2 CNN layers with stride 2 for time dimension for each layer, which results in time reduction rate 4 (TR4). The base frame rate is 10ms so the frame rate after sub-sampling layer is 40ms. The baseline model has 18 Conformer blocks in the speech encoder, each contains 512 hidden nodes, 8 attention heads, and 1024 feed-forward nodes. The prediction network has 2 LSTM layers with 320 embedding dimension and 1024 hidden nodes. The joint network is a single feed-forward layer with 512 nodes and the English word-piece size is 4K. The total number of parameters is 100M. The baseline model can produce high quality translation results. An additional baseline with time reduction rate 8 (TR8) by stacking 3 CNN layers with stride 2 for each layer is also built to further reduce decoding time. 

For CTC-GMM models, we set the CTC prediction branch at the $12^{th}$ layer of the speech encoder, and the shared encoder has 6 layers of Conformer blocks. The Conformer block setup is the same as the baseline conformer transducer. Therefore, the total number of parameters is still 100M. The CTC output reference is in German word pieces and the total vocabulary size is 2K. 


\section{Results}
\label{sec:res}
The evaluation results of the baseline and CTC-GMM models on FLEURS and CoVoST2 are shown in Table \ref{tab:result}. The translation quality is evaluated with both BLEU \cite{papineni2002bleu} and COMET \cite{rei2020comet} metrics. The real time factor (RTF) is measured on a single H100 GPU. 
The baseline conformer transducer ST model achieves 28.0 and 34.5 BLEU scores on FLEURS and CoVoST2, respectively. The RTF for the baseline model is 0.072. The baseline with time reduction 8 (TR8) gets lower RTF 0.049 as expected, but the accuracy also drops obviously with BLEU score 26.2 and 33.7 for FLEURS and CoVOST2 respectively. This indicates that reducing the frame rate by uniform downsampling may hurt the translation accuracy. The COMET scores also show the similar trend for these two models. For simplicity, we will solely use the BLEU score to discuss the translation quality of the various models.

\begin{table}
    \ninept
    \centering
    \begin{tabular}{|l|c|c|c|}
        \hline
        method & FLEURS & CoVoST2 & RTF \\
        \hline \hline
         baseline&  28.0/~0.774 & 34.5/~0.803 & 0.072 \\ \hline \hline        
         baseline TR8 &  26.2/~0.753 & 33.7/~0.792 & 0.049 \\ \hline \hline
         CTC average & 27.0/~0.769 & 34.7/~0.803  & 0.027\\ \hline
        \hspace{1mm} +sampling & 27.7/~0.776 & 34.8/~0.805  & 0.027\\ \hline
         \hspace{2.5mm} \textbf{+ MT text} & \textbf{29.7/~0.796} & \textbf{35.8/~0.814}  & 0.027\\ \hline
         \hspace{4mm} \textbf{\textit{+ big shared encoder}} & \textbf{\textit{31.9/~0.813}} & \textbf{\textit{36.7/~0.828}} & \textit{0.029}\\ \hline \hline 
         CTC attention +sampling & 28.0/~0.771 & 34.7/~0.805 & 0.024\\ \hline
         \hspace{1mm}  \textbf{+ MT text} & \textbf{29.1/~0.789} & \textbf{35.7/~0.813}  & 0.024 \\ \hline
        \hspace{2.5mm}  \textbf{\textit{+ big shared encoder}} & \textbf{\textit{31.0/~0.810}} & \textbf{\textit{37.0/~0.826}} & \textit{0.025} \\
        \hline \hline
         CTC discrete +sampling & 23.6/~0.715 & 31.0/~0.760 & 0.029 \\ \hline
        \hspace{1mm} \textbf{+ MT text} & \textbf{26.3/~0.745} & \textbf{32.4/~0.775}  & 0.029\\ \hline
         \hspace{2.5mm}  \textbf{\textit{+ big shared encoder}} & \textbf{\textit{28.9/~0.774}} & \textbf{\textit{34.4/~0.795}} & \textit{0.031}\\ \hline
         CTC discrete +sampling 
         & 23.6/~0.704 & 30.7/~0.761 & 0.024 \\ 
         blank removal &  &  &  \\ \hline         
         \hspace{1mm}  \textbf{+ MT text} & \textbf{26.0/~0.741} & \textbf{32.3/~0.773}  & 0.024\\ \hline
         \hspace{2.5mm}  \textbf{\textit{+ big shared encoder}} & \textbf{\textit{28.3/~0.768}} & \textbf{\textit{34.1/~0.793}} & \textit{0.026} \\ \hline 
    \end{tabular}
    \caption{The evaluation results of CTC-GMM models on FLEURS and CoVoST2, measured by BLEU/COMET scores (the higher the better), and RTF on an H100 GPU machine. The models using MT text data are \textbf{bolded}.    }
    \label{tab:result}
\end{table}

\subsection{CTC-GMM with average operation}

\begin{table*}[t!]
    \centering
    \begin{tabularx}{\textwidth}{|X|X|X|}

    \hline
         baseline ST & CTC-GMM ST & reference \\
         \hline \hline
          Apia is located on the capital of Samoa. & Apia is the capital of Samoa \textbf{located on the island of Upolu}. & Apia is the capital of Samoa. \textbf{The town is on the island of Upolu.} \\ \hline
          For example, students of \textbf{Bennet Gollum} design a website every year. & For example, students from \textbf{Bennet School North Carolina} design a website every year.   & For example, each year students from \textbf{Bennet School in North Carolina} design a website . \\ \hline
         We can use \textbf{Budapodes motorcycle taxis} to get around in Goma. & You can use \textbf{Boda Boda's motorcycle taxis} to get around in Goma. & You can use \textbf{boda-boda motorcycle taxi} to get around Goma. \\ \hline
         In this period of European history, \textbf{she stood rich and powerful Catholic Church}. & In this period of European history, \textbf{the rich and powerful Catholic Church was put to the test.} & During this period of European history,\textbf{ the Catholic Church, which had become rich and powerful, came under scrutiny.}\\ \hline
         \end{tabularx}
    \caption{The examples of how MT text data can help to boost the ST quality with CTC-GMM for translating German speech into English text. The major difference is \textbf{bolded}. }
    \label{tab:analysis}
\end{table*}

Without using the MT corpus, the CTC-guided compression with the average operation degrades the BLEU score on FLEURS while maintaining the performance on CoVoST2. We hypothesize that it is because the training/testing mismatch when using CTC maximum posteriors to predict the tokens. Usually, the CTC model predicts tokens much more accurate in training than in testing. To mitigate the issue, we propose to use the sampling strategy described in Section \ref{ssec:method} during training so that the CTC predicted tokens during training contains more errors, aligned with the testing situation better. As a result, the BLEU score on FLEURS goes up to 27.7 with the sampling strategy. The sampling is used as the default for all the remaining experiments. 

With the aid of MT text data, the model gains 1.7 and 1.3 BLEU score improvement on FLEURS and CoVoST2, respectively. Thanks to the CTC-guided compression, the input to the shared encoder is much shorter than the original sequence, resulting in significant RTF speed up. Additionally, we found that enlarging the shared encoder does not increase the decoding speed significantly given the CTC-compressed sequence. We, therefore, increase the size of shared encoder by adding 12 more Conformer blocks and get further improvements by 2.2 and 0.9 BLEU score on FLEURS and CoVoST2, respectively. The enlarged model has totally 150M parameters. Compared to the baseline conformer transducer model, the BLEU scores are improved from 28.0 and 34.5 to 31.9 and 36.7 on FLEURS and CoVoST2, respectively. The RTF is reduced from 0.072 to 0.029, a relative 59.7\% improvement on decoding efficiency. In other words, we obtain a fast and accurate streaming ST model with the proposed CTC-GMM model.

We showcase how the MT corpus improve the ST performance in Table \ref{tab:analysis}. 
Our baseline ST model was trained with pseudo labels generated from ASR transcriptions processed through an MT service. The ASR training set lacks rich entity coverage compared to the MT corpus. For instance, ``Boda Boda'' appears multiple times in the MT corpus but not in the ASR corpus. Incorporating MT training data significantly enhances entity accuracy as the first 3 examples. Additionally, MT data has correct grammar, unlike pseudo labels with errors, leading to more grammatically accurate translations as seen in the last example.

\subsection{CTC-GMM with attention operation}
The CTC-GMM model using CTC compression and sampling strategy obtains almost the same BLEU scores as the baseline ST model, with 28.0 and 34.7 BLEU scores on FLEURS and CoVoST2, respectively. With the help of MT data, the BLEU scores are improved to 29.1 and 35.7 on FLEURS and CoVoST2, respectively. Compared to its counter part with average operation, the CTC-GMM with attention operation does not show significant improvements on translation quality although it has an additional attention layer. However, it achieves a better RTF as 0.024 than the average operation (0.27) without the use of the big shared encoder. The reason is that the attention operation will merge blank frames into the non-blank frames as described in Section~\ref{ssec:method}, and thus further reduces the sequence length compared to the average operation. 

Finally, by increasing the shared encoder with additional 12 layers, the BLEU scores on FLEURS and CoVoST2 are boosted to 31.0 and 37.0, respectively. This only slightly increases the decoding RTF to 0.025. 

\subsection{CTC-GMM with discrete operation}

We have built two CTC-GMM recipes for the discrete option; one preserves the blank token, while the other excludes it. In both approaches, there is a notable decline in BLEU scores without supplemental MT text data; for instance, scores dropped from 28.0 to 23.6 on the FLEURS dataset. Introducing MT text data contributes to a significant increase in scores, yet they remain below the baseline. The scores are rose to 26.3 for the ``keeping blank'' approach and to 26.0 for ``removing blank'' approach on the FLEURS set, with little difference in ST quality. If we increase the shared encoder size to 18 layers, the ``keeping blank'' recipe gets better accuracy than the ``removing blank'' recipe, exemplified by BLEU scores of 28.9 versus 28.3 on FLEURS. Overall, the optimal outcomes for discrete operations achieved 28.9 on FLEURS and 34.4 on CoVoST2, comparable to or surpassing the baseline. In terms of decoding speed, the ``removing blank'' method outperforms the ``keeping blank'' due to its further reduction in acoustic sequence length. However, the ``keeping blank'' incurs a higher RTF than the average strategy, attributed to additional embedding processes.

\subsection{Frame rate evaluation}
As the RTF evaluation is machine dependent, we also list frame span of different models in Table \ref{tab:frame_rate} for reference. With the time reduction 4 and 8, the baseline models with TR 4 and TR 8 are with 40 ms and 80 ms frame rates, respectively. CTC compressor can significantly reduce the frame rate. On average, each frame after the CTC compressor spans 147 ms for both CTC average and discrete operations. CTC compressor with attention operation or removing blank in discrete operation can almost half the frame rate, with each frame  spanning 280 ms on average. The longer frame span results in the reduced RTF as shown in Table \ref{tab:result}.

\begin{table}
    \centering
    \begin{tabular}{|c|c|}
    \hline
        method & frame span (ms)\\ \hline
        baseline & 40\\ \hline
        baseline TR8 & 80\\ \hline
        CTC average & 147\\ \hline
        CTC attention & 280 \\ \hline
        CTC discrete & 147\\ \hline
        CTC discrete with blank removal & 280\\ \hline
    \end{tabular}
    \caption{Frame span of different models}
    \label{tab:frame_rate}
\end{table}

\subsection{Entity translation evaluation}
In Table \ref{tab:analysis}, we observed several examples that the CTC-GMM model can better translate the entity. Therefore in Table \ref{tab:entity}, we conducted a further evaluation by using an in-house German-to-English speech translation test set which has rich human-tagged entities. This test set contains 1942 utterances with totally 49454 words in source language transcription and 1700 entities. 

The baseline and the CTC-GMM (with the options of average, sampling, MT text, and big shared encoder) models are the ones in Table \ref{tab:result}. In addition to BLEU and COMET score improvement for this new in-house test set, there was a notable rise in entity recall rate from 62.6 to 68.4.

\begin{table}
    \centering
    \begin{tabular}{|c|c|c|}
    \hline
         & baseline & CTC-GMM \\ \hline
       BLEU & 31.7 & 33.7 \\ \hline
        COMET & 0.744 & 0.762\\ \hline
        entity recall & 62.6 & 68.4\\ \hline
    \end{tabular}
    \caption{Evaluation of an internal test set with rich entities}
    \label{tab:entity}
\end{table}

\section{Conclusions}
\label{sec:con}
In this paper, we presented CTC-GMM, the proposed method that uses CTC to compress the speech sequence to better align the text sequence. As a consequence, the MT training data can be employed to improve the quality of streaming ST models. In the proposed CTC-GMM, the sampling-based method was explored to predict the token for each frame instead of selecting the one with highest probabilities. Besides, several CTC compression methods: average, attention and discrete, were investigated, and the one with average operation gave the overall best ST quality improvements. We conduct the experiments by training ST models that translate German audio into English text. The BLEU scores were increased from 28.0 and 34.5 to 31.9 and 36.7 on FLEURS and CoVoST2, respectively. Since the compressed embedding sequence is much shorter, the decoding speed is significantly reduced accordingly. The RTF was decreased from 0.072 to 0.029 when evaluating with a single H100 GPU machine.

In conclusion, CTC-GMM could help to get fast and accurate speech translation model by using extra text translation data. Although we only reported the ST results using the German to English direction, We started to expand to a large amount of language pairs and are already seeing initial benefits which will be reported in the future. This approach is also applicable to other speech-related tasks like speech recognition, as well as different model structures such as AED.
\bibliographystyle{IEEEbib}
\bibliography{strings,refs}

\begin{thebibliography}{10}

\bibitem{sperber2020speech}
Matthias Sperber and Matthias Paulik,
\newblock ``Speech translation and the end-to-end promise: Taking stock of where we are,''
\newblock {\em arXiv preprint arXiv:2004.06358}, 2020.

\bibitem{Berard2016ST}
Alexandre Berard, Olivier Pietquin, Christophe Servan, and Laurent Besacier,
\newblock ``Listen and translate: A proof of concept for end-to-end speech-to-text translation,''
\newblock in {\em NIPS Workshop on End-to-end Learning for Speech and Audio Processing}, 2016.

\bibitem{weiss2017sequence}
Ron~J Weiss, Jan Chorowski, Navdeep Jaitly, Yonghui Wu, and Zhifeng Chen,
\newblock ``Sequence-to-sequence models can directly translate foreign speech,''
\newblock in {\em Proc. Interspeech}, 2017, pp. 2625--2629.

\bibitem{vila2018end}
Laura~Cross Vila, Carlos Escolano, Jos{\'e}~AR Fonollosa, and Marta~R Costa-Jussa,
\newblock ``End-to-end speech translation with the transformer.,''
\newblock in {\em Proc. Interspeech}, 2018, pp. 60--63.

\bibitem{radford2023robust}
Alec Radford, Jong~Wook Kim, Tao Xu, Greg Brockman, Christine McLeavey, and Ilya Sutskever,
\newblock ``Robust speech recognition via large-scale weak supervision,''
\newblock in {\em International Conference on Machine Learning}. PMLR, 2023, pp. 28492--28518.

\bibitem{cho2014learning}
Kyunghyun Cho, Bart Van~Merri{\"e}nboer, Caglar Gulcehre, Dzmitry Bahdanau, Fethi Bougares, Holger Schwenk, and Yoshua Bengio,
\newblock ``Learning phrase representations using {RNN} encoder-decoder for statistical machine translation,''
\newblock {\em arXiv preprint arXiv:1406.1078}, 2014.

\bibitem{Attention-bahdanau2014}
Dzmitry Bahdanau, Kyunghyun Cho, and Yoshua Bengio,
\newblock ``Neural machine translation by jointly learning to align and translate,''
\newblock {\em arXiv preprint arXiv:1409.0473}, 2014.

\bibitem{ma2021streaming}
Xutai Ma, Yongqiang Wang, Mohammad~Javad Dousti, Philipp Koehn, and Juan Pino,
\newblock ``Streaming simultaneous speech translation with augmented memory transformer,''
\newblock in {\em Proc. ICASSP}. IEEE, 2021, pp. 7523--7527.

\bibitem{chiu2018monotonic}
Chung-Cheng Chiu and Colin Raffel,
\newblock ``Monotonic chunkwise attention,''
\newblock in {\em International Conference on Learning Representations}, 2018.

\bibitem{arivazhagan2019monotonic}
Naveen Arivazhagan, Colin Cherry, Wolfgang Macherey, Chung-Cheng Chiu, Semih Yavuz, Ruoming Pang, Wei Li, and Colin Raffel,
\newblock ``Monotonic infinite lookback attention for simultaneous machine translation,''
\newblock in {\em Proceedings of the Annual Meeting of the Association for Computational Linguistics}, 2019, pp. 1313--1323.

\bibitem{ma2019monotonic}
Xutai Ma, Juan~Miguel Pino, James Cross, Liezl Puzon, and Jiatao Gu,
\newblock ``Monotonic multihead attention,''
\newblock in {\em Proceedings of International Conference on Learning Representations}, 2019.

\bibitem{barrault2023seamless}
Lo{\"\i}c Barrault, Yu-An Chung, Mariano~Coria Meglioli, et~al.,
\newblock ``Seamless: Multilingual expressive and streaming speech translation,''
\newblock {\em arXiv preprint arXiv:2312.05187}, 2023.

\bibitem{sainath2020streaming}
Tara~N Sainath, Yanzhang He, Bo~Li, Arun Narayanan, et~al.,
\newblock ``A streaming on-device end-to-end model surpassing server-side conventional model quality and latency,''
\newblock in {\em Proc. ICASSP}. IEEE, 2020, pp. 6059--6063.

\bibitem{Li2020Developing}
Jinyu Li, Rui Zhao, Zhong Meng, Yanqing Liu, et~al.,
\newblock ``Developing {RNN-T} models surpassing high-performance hybrid models with customization capability,''
\newblock in {\em Proc. Interspeech}, 2020, pp. 3590--3594.

\bibitem{Graves-RNNSeqTransduction}
Alex Graves,
\newblock ``Sequence transduction with recurrent neural networks,''
\newblock {\em arXiv preprint arXiv:1211.3711}, 2012.

\bibitem{li2022recent}
Jinyu Li,
\newblock ``Recent advances in end-to-end automatic speech recognition,''
\newblock {\em APSIPA Transactions on Signal and Information Processing}, vol. 11, no. 1, 2022.

\bibitem{prabhavalkar2023end}
Rohit Prabhavalkar, Takaaki Hori, Tara~N Sainath, Ralf Schl{\"u}ter, and Shinji Watanabe,
\newblock ``End-to-end speech recognition: A survey,''
\newblock {\em IEEE/ACM Transactions on Audio, Speech, and Language Processing}, 2023.

\bibitem{xue2022large}
Jian Xue, Peidong. Wang, Jinyu Li, Matt Post, and Yashesh Gaur,
\newblock ``Large-scale streaming end-to-end speech translation with neural transducers,''
\newblock in {\em Proc. Interspeech}, 2022, pp. 3263--3267.

\bibitem{xue2023weakly}
Jian Xue, Peidong Wang, Jinyu Li, and Eric Sun,
\newblock ``A weakly-supervised streaming multilingual speech model with truly zero-shot capability,''
\newblock in {\em Proc. ASRU}. IEEE, 2023, pp. 1--7.

\bibitem{3pAPI}
Microsoft,
\newblock ``Announcing video translation \& speech translation {API} enhancements,'' \url{https://techcommunity.microsoft.com/t5/ai-azure-ai-services-blog/announcing-video-translation-amp-speech-translation-api/ba-p/4148007}, 2024.

\bibitem{PCtranslation}
Microsoft,
\newblock ``Introducing {Copilot+ PCs},'' \url{https://blogs.microsoft.com/blog/2024/05/20/introducing-copilot-pcs/}, 2024.

\bibitem{jia2019st}
Ye~Jia, Melvin Johnson, Wolfgang Macherey, Ron~J Weiss, Yuan Cao, Chung-Cheng Chiu, Naveen Ari, Stella Laurenzo, and Yonghui Wu,
\newblock ``Leveraging weakly supervised data to improve end-to-end speech-to-text translation,''
\newblock in {\em Proc. ICASSP}. IEEE, 2019, pp. 7180--7184.

\bibitem{gaido2020end}
Marco Gaido, Mattia~A Di~Gangi, Matteo Negri, and Marco Turchi,
\newblock ``End-to-end speech-translation with knowledge distillation,''
\newblock in {\em Proceedings of the International Conference on Spoken Language Translation}, 2020, pp. 80--88.

\bibitem{graves2006connectionist}
Alex Graves, Santiago Fern{\'a}ndez, Faustino Gomez, and J{\"u}rgen Schmidhuber,
\newblock ``Connectionist temporal classification: labelling unsegmented sequence data with recurrent neural networks,''
\newblock in {\em Proceedings of the 23rd international conference on Machine learning}. ACM, 2006, pp. 369--376.

\bibitem{fleurs2022arxiv}
Alexis Conneau, Min Ma, Simran Khanuja, Yu~Zhang, et~al.,
\newblock ``{FLEURS}: Few-shot learning evaluation of universal representations of speech,''
\newblock {\em arXiv preprint arXiv:2205.12446}, 2022.

\bibitem{wang2020covost}
Changhan Wang, Anne Wu, and Juan Pino,
\newblock ``Covost 2 and massively multilingual speech-to-text translation,''
\newblock {\em arXiv preprint arXiv:2007.10310}, 2020.

\bibitem{gaido2021ctc}
Marco Gaido, Mauro Cettolo, Matteo Negri, and Marco Turchi,
\newblock ``{CTC}-based compression for direct speech translation,''
\newblock {\em arXiv preprint arXiv:2102.01578}, 2021.

\bibitem{ruchao2023sampling}
Ruchao Fan, Wei Chu, Peng Chang, and Alwan Abeer,
\newblock ``A {CTC} alignment-based non-autoregressive transformer for end-to-end automatic speech recognition,''
\newblock {\em IEEE/ACM Transactions on Audio, Speech, and Language Processing}, 2023.

\bibitem{ao2022speecht5}
Junyi Ao, Rui Wang, Long Zhou, Chengyi Wang, et~al.,
\newblock ``{SpeechT5}: Unified-modal encoder-decoder pre-training for spoken language processing,''
\newblock in {\em Proc. ACL}, 2022, pp. 5723--5738.

\bibitem{zhang2024speechlm}
Ziqiang Zhang, Sanyuan Chen, Long Zhou, Yu~Wu, Shuo Ren, Shujie Liu, Zhuoyuan Yao, Xun Gong, Lirong Dai, Jinyu Li, et~al.,
\newblock ``{SpeechLM}: Enhanced speech pre-training with unpaired textual data,''
\newblock {\em IEEE/ACM Transactions on Audio, Speech, and Language Processing}, 2024.

\bibitem{zhao2022Madaptor}
Jinming Zhao, Hao Yang, Gholamreza Haffari, and Ehsan Shareghi,
\newblock ``M-adapter: Modality adaptation for end-to-end speech-to-text translation,''
\newblock in {\em Proc. Interspeech}, 2022, pp. 111--115.

\bibitem{chen22MAESTRO}
Zhehuai Chen, Yu~Zhang, Andrew Rosenberg, Bhuvana Ramabhadran, Pedro~J. Moreno, Ankur Bapna, and Heiga Zen,
\newblock ``{MAESTRO}: Matched speech text representations through modality matching,''
\newblock in {\em Proc. Interspeech}, 2022, pp. 4093--4097.

\bibitem{zhao2021addressing}
Rui Zhao, Jian Xue, Jinyu Li, Wenning Wei, Lei He, and Yifan Gong,
\newblock ``On addressing practical challenges for {RNN}-transducer,''
\newblock in {\em Proc. ASRU}. IEEE, 2021, pp. 526--533.

\bibitem{wang2023neural}
Chengyi Wang, Sanyuan Chen, Yu~Wu, Ziqiang Zhang, et~al.,
\newblock ``Neural codec language models are zero-shot text to speech synthesizers,''
\newblock {\em arXiv preprint arXiv:2301.02111}, 2023.

\bibitem{borsos2023soundstorm}
Zal{\'a}n Borsos, Matt Sharifi, Damien Vincent, Eugene Kharitonov, Neil Zeghidour, and Marco Tagliasacchi,
\newblock ``Soundstorm: Efficient parallel audio generation,''
\newblock {\em arXiv preprint arXiv:2305.09636}, 2023.

\bibitem{ju2024naturalspeech}
Zeqian Ju, Yuancheng Wang, Kai Shen, Xu~Tan, Detai Xin, et~al.,
\newblock ``Naturalspeech 3: Zero-shot speech synthesis with factorized codec and diffusion models,''
\newblock in {\em Proc. ICML}, June 2024.

\bibitem{le2024voicebox}
Matthew Le, Apoorv Vyas, Bowen Shi, Brian Karrer, et~al.,
\newblock ``Voicebox: Text-guided multilingual universal speech generation at scale,''
\newblock {\em Advances in neural information processing systems}, vol. 36, 2024.

\bibitem{burchi2021efficient}
Maxime Burchi and Valentin Vielzeuf,
\newblock ``Efficient conformer: Progressive downsampling and grouped attention for automatic speech recognition,''
\newblock in {\em Proc. ASRU}. IEEE, 2021, pp. 8--15.

\bibitem{li23Accelerating}
Yuang Li, Yu~Wu, Jinyu Li, and Shujie Liu,
\newblock ``Accelerating transducers through adjacent token merging,''
\newblock in {\em Proc. Interspeech}, 2023, pp. 1379--1383.

\bibitem{prabhavalkar2024extreme}
Rohit Prabhavalkar, Zhong Meng, Weiran Wang, et~al.,
\newblock ``Extreme encoder output frame rate reduction: Improving computational latencies of large end-to-end models,''
\newblock in {\em Proc. ICASSP}. IEEE, 2024, pp. 11816--11820.

\bibitem{li2012improving}
Jinyu Li, Dong Yu, Jui-Ting Huang, and Yifan Gong,
\newblock ``Improving wideband speech recognition using mixed-bandwidth training data in {CD-DNN-HMM},''
\newblock in {\em Proc. SLT}. IEEE, 2012, pp. 131--136.

\bibitem{gulati2020conformer}
Anmol Gulati, James Qin, Chung-Cheng Chiu, Niki Parmar, Yu~Zhang, et~al.,
\newblock ``Conformer: Convolution-augmented transformer for speech recognition,''
\newblock in {\em Proc. Interspeech}, 2020, pp. 5036--5040.

\bibitem{chen2021developing}
Xie Chen, Yu~Wu, Zhenghao Wang, Shujie Liu, and Jinyu Li,
\newblock ``Developing real-time streaming transformer transducer for speech recognition on large-scale dataset,''
\newblock in {\em Proc. ICASSP}. IEEE, 2021, pp. 5904--5908.

\bibitem{papineni2002bleu}
Kishore Papineni, Salim Roukos, Todd Ward, and Wei-Jing Zhu,
\newblock ``{BLEU}: a method for automatic evaluation of machine translation,''
\newblock in {\em Proceedings of the 40th annual meeting of the Association for Computational Linguistics}, 2002, pp. 311--318.

\bibitem{rei2020comet}
Ricardo Rei, Craig Stewart, Ana~C Farinha, and Alon Lavie,
\newblock ``{COMET}: A neural framework for mt evaluation,''
\newblock {\em arXiv preprint arXiv:2009.09025}, 2020.

\end{thebibliography}

\end{document}